\pdfoutput=1

\documentclass{article}
\usepackage{spconf}
\usepackage{graphicx}
\usepackage{amsmath,amsfonts,amssymb}
\usepackage[caption=false,font=footnotesize]{subfig}
\usepackage{epstopdf}
\usepackage{algorithm}
\usepackage{algpseudocode}
\usepackage{color}
\usepackage{booktabs}
\usepackage{multirow}
\usepackage{comment}

 \usepackage{color}
\DeclareGraphicsExtensions{.eps,.pdf}	
\graphicspath{{./figures/}}

\floatname{algorithm}{Method}

\title{THREE DIMENSIONAL BLIND IMAGE DECONVOLUTION FOR FLUORESCENCE \\ MICROSCOPY USING GENERATIVE ADVERSARIAL NETWORKS}

\name{Soonam Lee$^{\star}$ \quad Shuo Han$^{\star}$ \quad Paul Salama$^{\dagger}$ \quad Kenneth W. Dunn$^{\ddagger}$ \quad Edward J. Delp$^{\star}$ \thanks{This work was partially supported by a George M. O'Brien Award from the National Institutes of Health under grant NIH/NIDDK P30 DK079312
and the endowment of the Charles William Harrison Distinguished Professorship at Purdue University.}}
\address{\normalsize \parbox{2.5in}{\centering
		$^{\star}$Video and Image Processing Laboratory\\
		School of Electrical and Computer Engineering\\
		Purdue University\\
		West Lafayette, Indiana}
	\parbox{2.25in}{\centering
		$^{\dagger}$Department of Electrical and \\Computer Engineering\\
		Indiana University\\
		Indianapolis, Indiana}
	\parbox{2.25in}{\centering
		$^{\ddagger}$Division of Nephrology\\School of Medicine\\
		Indiana University\\
		Indianapolis, Indiana}
\\[-1.75ex]
}

\begin{document}
\ninept
\maketitle
\nointerlineskip

\vspace{-0.1in}
\begin{abstract}
Due to image blurring image deconvolution is often used for studying biological structures in fluorescence microscopy. Fluorescence microscopy image volumes inherently suffer from intensity inhomogeneity, blur, and are corrupted by various types of noise which exacerbate image quality at deeper tissue depth. Therefore, quantitative analysis of fluorescence microscopy in deeper tissue still remains a challenge. This paper presents a three dimensional blind image deconvolution method for fluorescence microscopy using $3$-way spatially constrained cycle-consistent adversarial networks. The restored volumes of the proposed deconvolution method and other well-known deconvolution methods, denoising methods, and an inhomogeneity correction method are visually and numerically evaluated. Experimental results indicate that the proposed method can restore and improve the quality of blurred and noisy deep depth microscopy image visually and quantitatively. 
\end{abstract}
\vspace{-0.05in}

\vspace{-0.1in}
\section{INTRODUCTION}
\label{sec:intro}
\vspace{-0.1in}

Fluorescence microscopy is a modality that allows imaging of subcellular structures from live specimens \cite{bib:Piston1999, bib:Vonesch2006}. During this image acquisition process, large datasets of 3D microscopy image volumes are generated.
The quantitative analysis of the fluorescence microscopy volume is hampered by light diffraction, distortion created by lens aberrations in different directions, complex variation of biological structures \cite{bib:Murphy2012}. The image acquisition process can be typically modeled as the convolution of the observed objects with a 3D point spread function (PSF) followed by degradation from noise such as Poisson noise and Gaussian noise \cite{bib:Yang2015}. 
These limitations result in anisotropic, inhomogeneous background, blurry (out-of-focus), and noisy image volume with poor edge details which aggravate the image quality poorer in depth \cite{bib:SLee2018a}.

There has been various approaches to improve 3D fluorescence microscopy images quality. One popular approach is known as image deconvolution which ``inverts'' the convolution process to restore the original microscopy image volume \cite{bib:Sarder2006}. 
Richard-Lucy (RL) deconvolution \cite{bib:Richard1972, bib:Lucy1974} which maximizes the likelihood distribution based on a Poisson noise assumption for confocal microscopy was proposed. This RL deconvolution was further extended in \cite{bib:Dey2006} that incorporated the total variation as a regularization term in the cost function. Since the PSF is usually not known, blind deconvolution which estimates the PSF and the original image simultaneously is favorable \cite{bib:Shajkofci2018}. Blind deconvolution using RL deconvolution was described in \cite{bib:Fish1995}. A pupil model for the PSF was presented in \cite{bib:Soulez2012} and the PSF was estimated using machine learning approaches in \cite{bib:Kenig2010}. Sparse coding to learn 2D features for coarse resolution along the depth axis to mitigate anisotropic issues was presented in \cite{bib:Soulez2014}.

Another approach to achieve better image quality stems from image denoising research. 
One example is Poisson noise removal using a combination of the Haar wavelet and the linear expansion of thresholds (PURE-LET) proposed in \cite{bib:PureDenoise}. This PURE-LET approach was extended further to 3D widefield microscopy in \cite{bib:JLi2017}. Additionally, a denoising and deblurring method for Poisson noise corrupted data using variance stabilizing transforms (VST) was described in \cite{bib:Azzari2017}. Meanwhile, a 3D inhomogeneity correction method that combines 3D active contours segmentation was presented in \cite{bib:SLee2017}. 

Convolutional neural network (CNN) has been popular to address various problems in medical image analysis and computer vision  such as image denoising, image segmentation, and image registration \cite{bib:Litjens2017}.
There are few papers that focus on image deconvolution in fluorescence microscopy using CNNs. 
One example is an anisotropic fluorescence microscopy restoration method using a CNN \cite{bib:Weigert2017}. Later, semi-blind spatially-variant deconvolution in optical microscopy with a local PSF using a CNN was described in \cite{bib:Shajkofci2018}. 
More recently, generative adversarial networks has gradually gained interest in medical imaging  especially for medical image analysis \cite{bib:Yi2018}. One of the useful architectures for medical image is a cycle-consistent adversarial networks (CycleGAN) \cite{bib:CycleGAN} which learns image-to-image translation without having paired images (actual groundtruth images). This CycleGAN was utilized for CT denoising by an analogy of mapping low dose phase images to high dose phase images to improve image quality \cite{bib:Kang2018}. 
Additionally, this CycleGAN was further extended by \cite{bib:CFu2018} incorporating a spatial constrained term to minimize misalignment between synthetically generated binary volume and corresponding synthetic microscopy volume to achieve better segmentation results. 


In this paper, we present a new approach to restore various biological structures in 3D microscopy images in deeper tissue without knowing the 3D PSF using a spatially constrained CycleGAN (SpCycleGAN) \cite{bib:CFu2018}. We train and inference the SpCycleGAN in three directions along with $xy$, $yz$, and $xz$ sections ($3$-Way SpCycleGAN) to incorporate 3D information inspired by \cite{bib:Roth2014, bib:SLee2018b}. These restored $3$-way microscopy images are then averaged and evaluated with three different image quality metrics. Our datasets consist of Hoechst $33342$ labeled nuclei and a phalloidin labeled filamentous actin collected from a rat kidney using two-photon microscopy. 
The goal is to restore blurred and noisy 3D microscopy images to the level of well-defined images so that the deeper depth tissues can be used for biological study.

\vspace{-0.2in}
\section{PROPOSED METHOD}
\label{sec:method}
\vspace{-0.15in}

\begin{figure}[htbp!]
	\centering
	\includegraphics[width=0.5\textwidth]{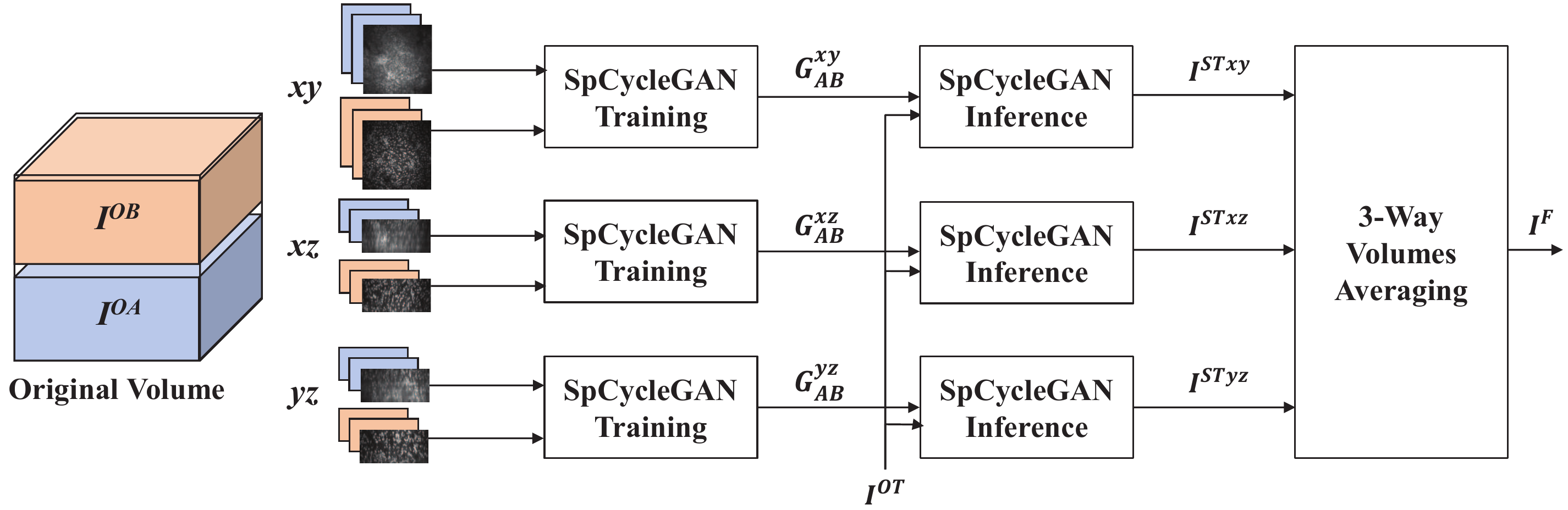}
	\vspace{-0.3in}
	\caption{Block diagram of the proposed deconvolution method using a $3$-Way SpCycleGAN} \label{fig:blockdiagram}
\end{figure}
\vspace{-0.15in}

Figure \ref{fig:blockdiagram} shows a block diagram of the proposed 3D images deconvolution method. 
We denote $I$ as a 3D image volume of size $X \times Y \times Z$. Note that $I_{z_p}$ is a $xy$ section with $p^\text{th}$ focal plane along the $z$-direction in a volume, where $p \in \{1, \dots, Z\}$. Similarly, $I_{y_q}$ is a $xz$ section with $q^\text{th}$ focal plane along $y$-direction, where $q \in \{1, \dots, Y\}$, and $I_{x_r}$ is a $yz$ section with $r^\text{th}$ focal plane along $x$-direction, where $r \in \{1, \dots, X\}$. 
In addition, let $I_{\left(r_i:r_f,q_i:q_f,p_i:p_f\right)}$ be a subvolume of $I$, whose $x$-coordinate is $r_i \leq x \leq r_f$, $y$-coordinate is $q_i \leq y \leq q_f$, and $z$-coordinate is $p_i \leq z \leq p_f$.
For example, $I_{\left(241:272,241:272,131:162\right)}$ is a subvolume of $I$ where the subvolume is cropped between $241^\text{st}$ slice and $272^\text{nd}$ slice in $x$-direction, between $241^\text{st}$ slice and $272^\text{nd}$ slice in $y$-direction, and between $131^\text{st}$ slice and $162^\text{nd}$ slice in $z$-direction.

As shown in Figure \ref{fig:blockdiagram}, we divide an original florescence microscopy volume denoted as $I^{O}$ into two subvolumes such as an out-of-focus and noisy subvolume and a well-defined subvolume denoted as $I^{OA}$ and $I^{OB}$, respectively. 
In particular, we choose $I^{OA}$ from deep sections and $I^{OB}$ from shallow sections since shallow sections of fluorescence microscopy volumes typically have a better image quality than deep sections.
These two volumes are sliced in the $z$-, $y$-, and $x$-direction to form the $xy$, $xz$, and $yz$ sections of the images. 
Then, the $xy$ sections from $I^{OA}$ and $I^{OB}$ are used for the training of the SpCycleGAN \cite{bib:CFu2018} to obtain the trained generative network denoted as $G_{AB}^{xy}$. 
Similarly doing this with the $xz$ sections and the $yz$ sections, trained generative networks $G_{AB}^{xz}$ and $G_{AB}^{yz}$ are obtained.
These generative networks are used for inference with a test volume denoted as $I^{OT}$ in the $xy$, $xz$, and $yz$ sections. 
Next, these synthetically generated results by the SpCycleGAN inference are stacked with $z$-, $y$-, and $x$-direction to form 3D volumes denoted as $I^{STxy}$, $I^{STxz}$, and $I^{STyz}$, respectively. 
Finally, we obtain the final volume $I^F$ by voxelwise weighted averaging of these volumes.

\vspace{-0.05in}
\subsection{Spatially Constrained CycleGAN (SpCycleGAN)}
\vspace{-0.05in}
This SpCycleGAN was introduced in our previous work \cite{bib:CFu2018} which extended the CycleGAN \cite{bib:CycleGAN} by adding one more term to the loss function and introducing an additional generative network. 
The SpCycleGAN was used for generating synthetic microscopy volumes from a synthetic binary volume. 
One problem with the CycleGAN is that the generated images sometimes are misaligned with the input images. 
We added a spatial constrained term ($\mathcal{L}_{spatial}$) to the loss function and minimize the loss function together with the two original GAN losses ($\mathcal{L}_{\text{GAN}}$) and the cycle consistent loss ($\mathcal{L}_{cyc}$) as:

\vspace{-0.2in}
\begin{align}\label{eq:LSpCycleGAN}
\mathcal{L}(G_{AB}, G_{BA}, H, D_A, D_B) &= \mathcal{L}_{\text{GAN}}(G_{AB},D_B,I^{OA}, I^{OB}) \nonumber \\
&+ \mathcal{L}_{\text{GAN}}(G_{BA},D_A,I^{OB},I^{OA}) \nonumber \\ 
&+ \lambda_1 \mathcal{L}_{cyc}(G_{AB},G_{BA}, I^{OA},I^{OB}) \nonumber \\
&+ \lambda_2 \mathcal{L}_{spatial}(G_{AB},H,I^{OA},I^{OB}) 
\end{align}
\vspace{-0.1in}
where
\\
\resizebox{1.00\hsize}{!}
{ \begin{minipage}{\linewidth}
\begin{align}
\mathcal{L}_{\text{GAN}}(G_{AB},D_B,I^{OA}, I^{OB}) &= \mathbb{E}_{I^{OB}}[\text{log}({D_B (I^{OB})})] \nonumber \\
&+ \mathbb{E}_{I^{OA}}[\text{log}(1 - {D_B (G_{AB} (I^{OA})))}] \nonumber
\end{align}
\end{minipage} 
} 
\\
\resizebox{1.00\hsize}{!}
{ \begin{minipage}{\linewidth}
\begin{align}
\mathcal{L}_{\text{GAN}}(G_{BA},D_A,I^{OB},I^{OA}) &= \mathbb{E}_{I^{OA}}[\text{log}({D_A (I^{OA})})] \nonumber \\
&+ \mathbb{E}_{I^{OB}}[\text{log}(1 - {D_A (G_{BA} (I^{OB})))}] \nonumber  
\end{align}
\end{minipage} 
} 
\\
\resizebox{1.00\hsize}{!}
{ \begin{minipage}{\linewidth}
\begin{align}
\mathcal{L}_{cyc}(G_{AB},G_{BA},I^{OA},I^{OB}) &= \mathbb{E}_{I^{OA}}[|| G_{BA}(G_{AB}(I^{OA})) - I^{OA} ||_1] \nonumber \\
&+ \mathbb{E}_{I^{OB}}[|| G_{AB}(G_{BA}(I^{OB})) - I^{OB} ||_1] \nonumber
\end{align}
\end{minipage} 
} 
\\
\resizebox{1.00\hsize}{!}
{ \begin{minipage}{\linewidth}
\begin{align}
\mathcal{L}_{spatial}(G_{AB}, H, I^{OA},I^{OB}) &= \mathbb{E}_{I^{OA}}[|| H(G_{AB}(I^{OA})) - I^{OA} ||_2] \nonumber.
\end{align}
\end{minipage} 
} \vspace{0.01in}
\\
\noindent 
Note that $\lambda_1$ and $\lambda_2$ are the controllable coefficients for $\mathcal{L}_{cyc}$ and $\mathcal{L}_{spatial}$. Also, $|| \cdot ||_1$ and $|| \cdot ||_2$ represent $L_1$ and $L_2$ norms, respectively. 
The generative model $G_{AB}$ transfers $I^{OA}$ to $I^{OB}$ and the generative model $G_{BA}$ transfers $I^{OB}$ to $I^{OA}$. 
Similarly, the discriminative model $D_A$ and $D_B$ distinguish between $I^{OA}$ and $G_{BA}(I^{OB})$ and between $I^{OB}$ and $G_{AB}(I^{OA})$. 
In particular, $G_{AB}(\cdot)$ is a transfer function using model $G_{AB}$ and $G_{BA}(\cdot)$ is another transfer function using model $G_{BA}$. 
For example, $G_{AB}(I^{OA})$ is a synthetically restored volume generated by model $G_{AB}$ using a blurred and noisy volume. Also, $G_{BA}(I^{OB})$ is a synthetically generating blurred and noisy volume by model $G_{BA}$ using a well-defined volume. Additionally, another generative model $H$ takes $G_{AB}(I^{OA})$ as an input to generate a synthetically blurred and noisy volume $H(G_{AB}(I^{OA}))$ using synthetically restored volume. This generative model $H$ minimizes $\mathcal{L}_2$ loss between $I_{OA}$ and $H(G_{AB}(I^{OA}))$.

\begin{figure*}[htb!]
	\centering
	\subfloat[]
		 {\label{fig:WSM_B_testA}\includegraphics[width=0.14\textwidth]{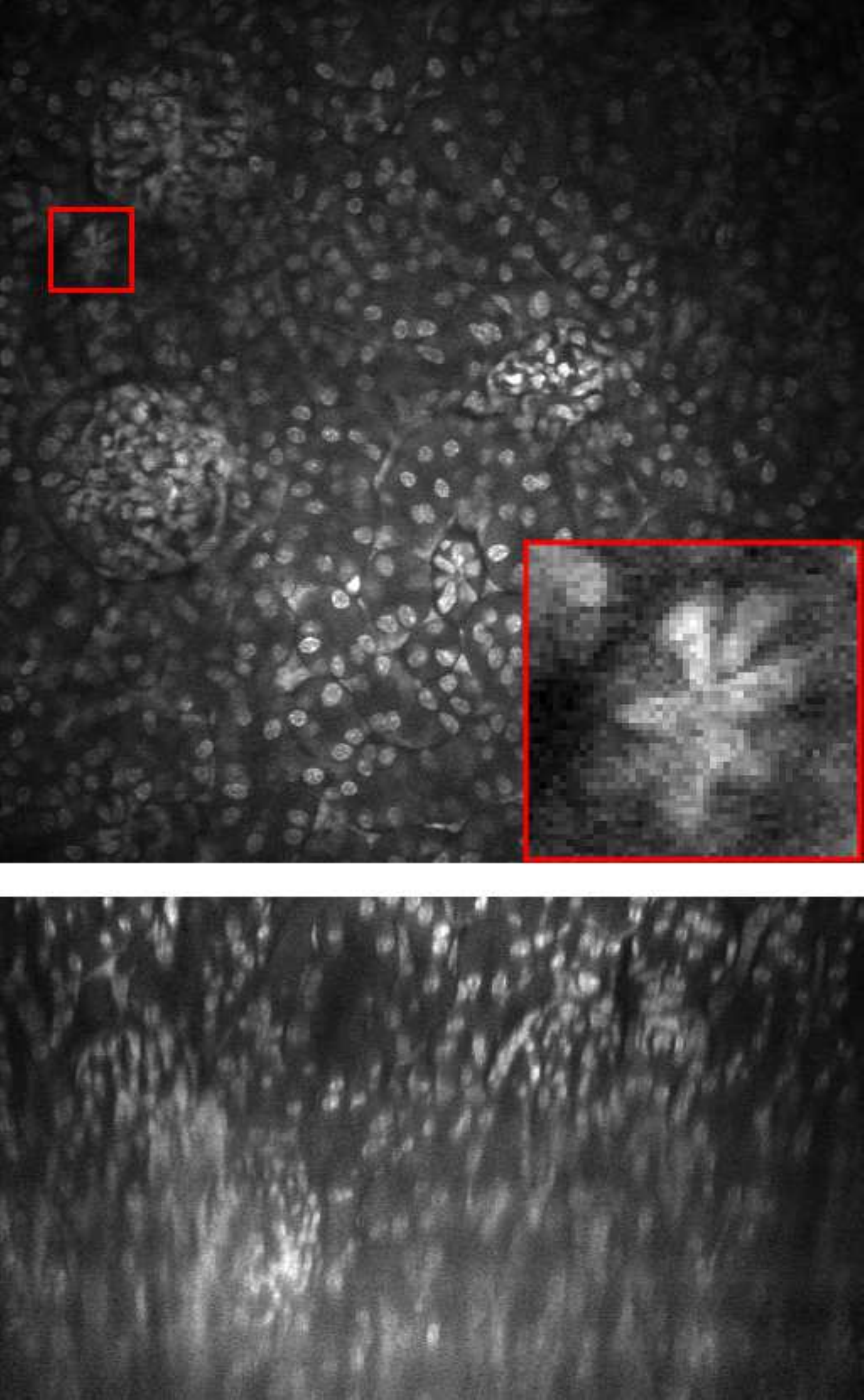}}
	\,
	\subfloat[]
		 {\label{fig:WSM_B_PSFSynRL}\includegraphics[width=0.14\textwidth]{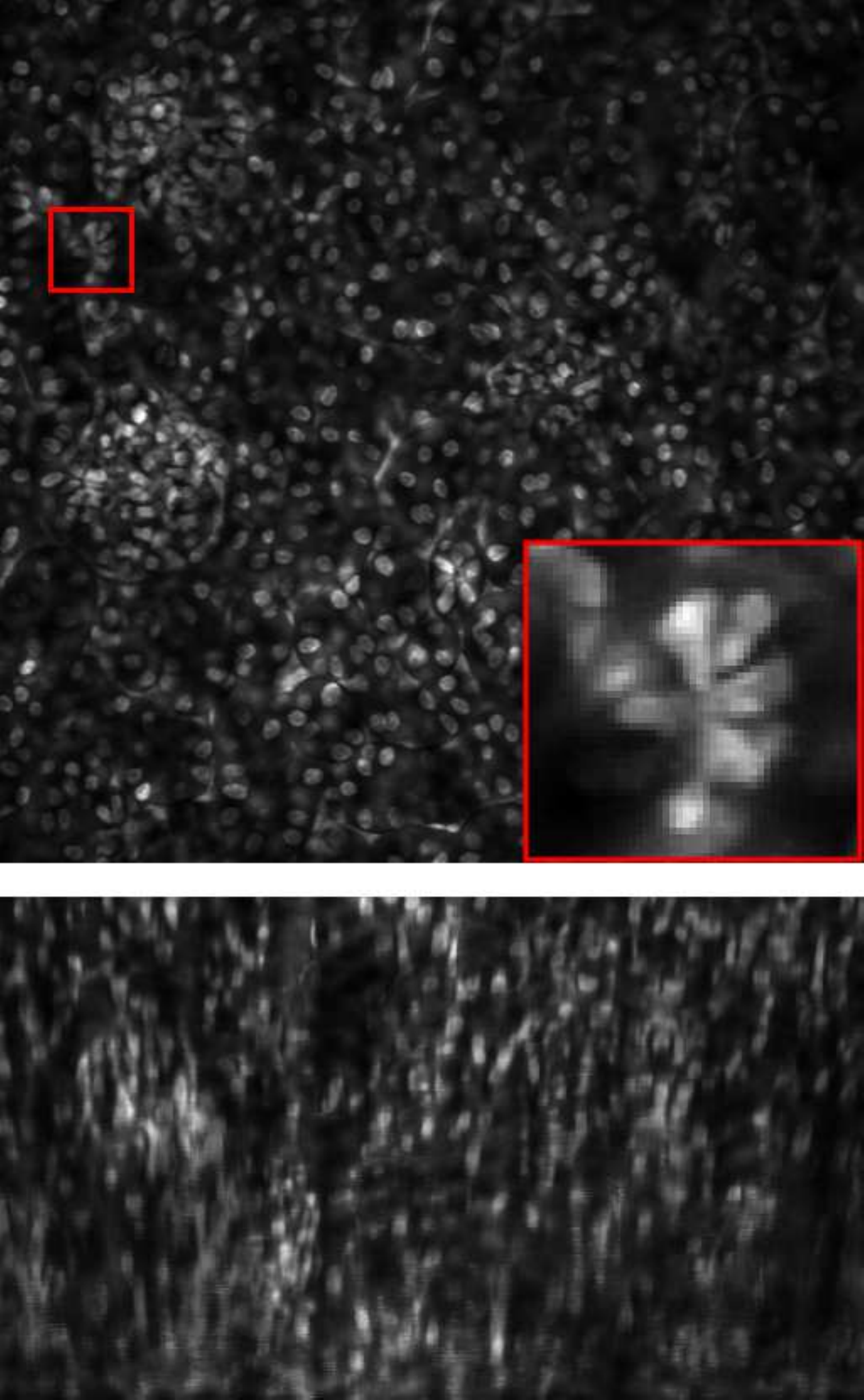}}
	\,
	\subfloat[]
		 {\label{fig:WSM_B_EpiDEMIC}\includegraphics[width=0.14\textwidth]{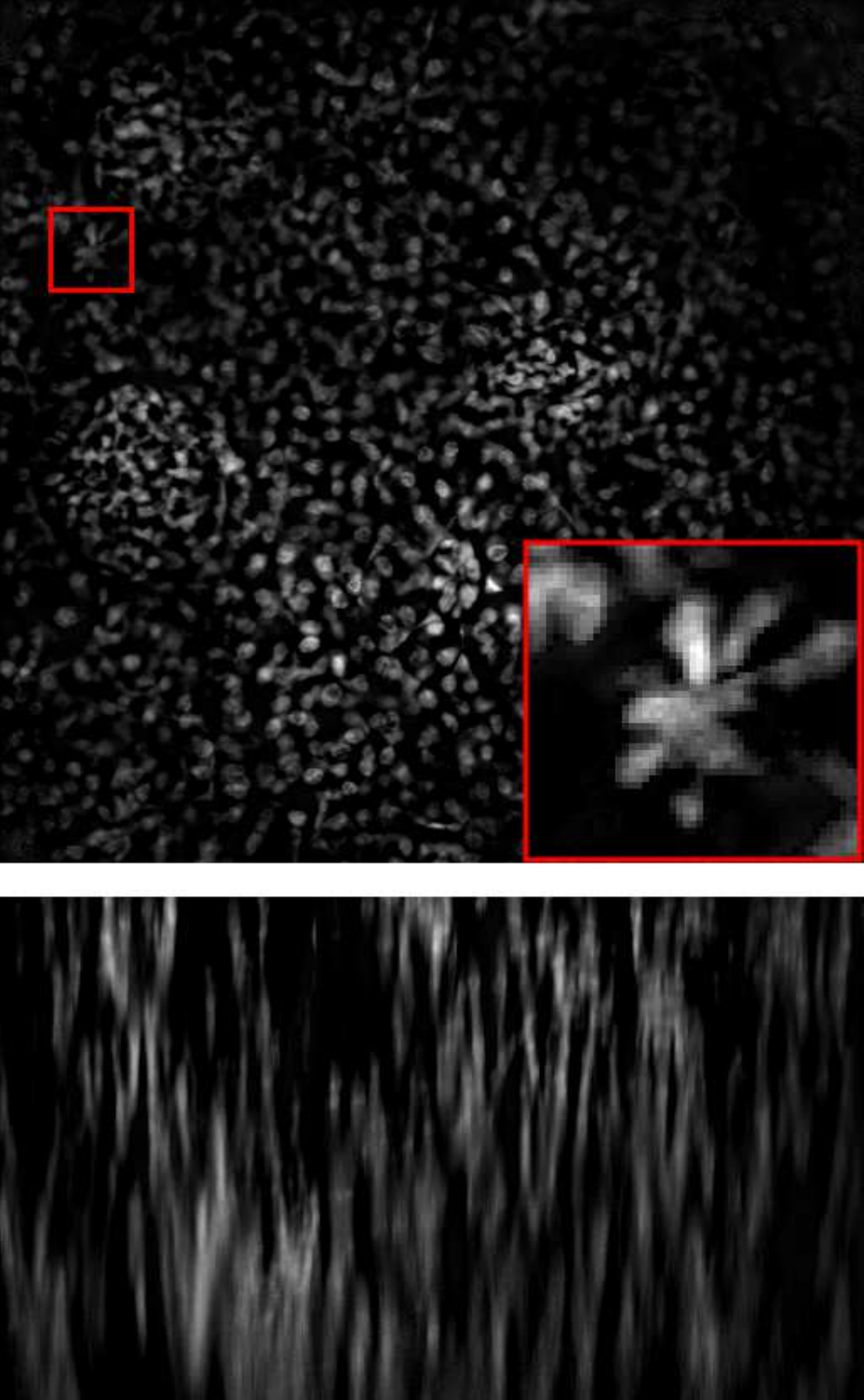}}
	\,
	\subfloat[]
		 {\label{fig:WSM_B_PureDenoise}\includegraphics[width=0.14\textwidth]{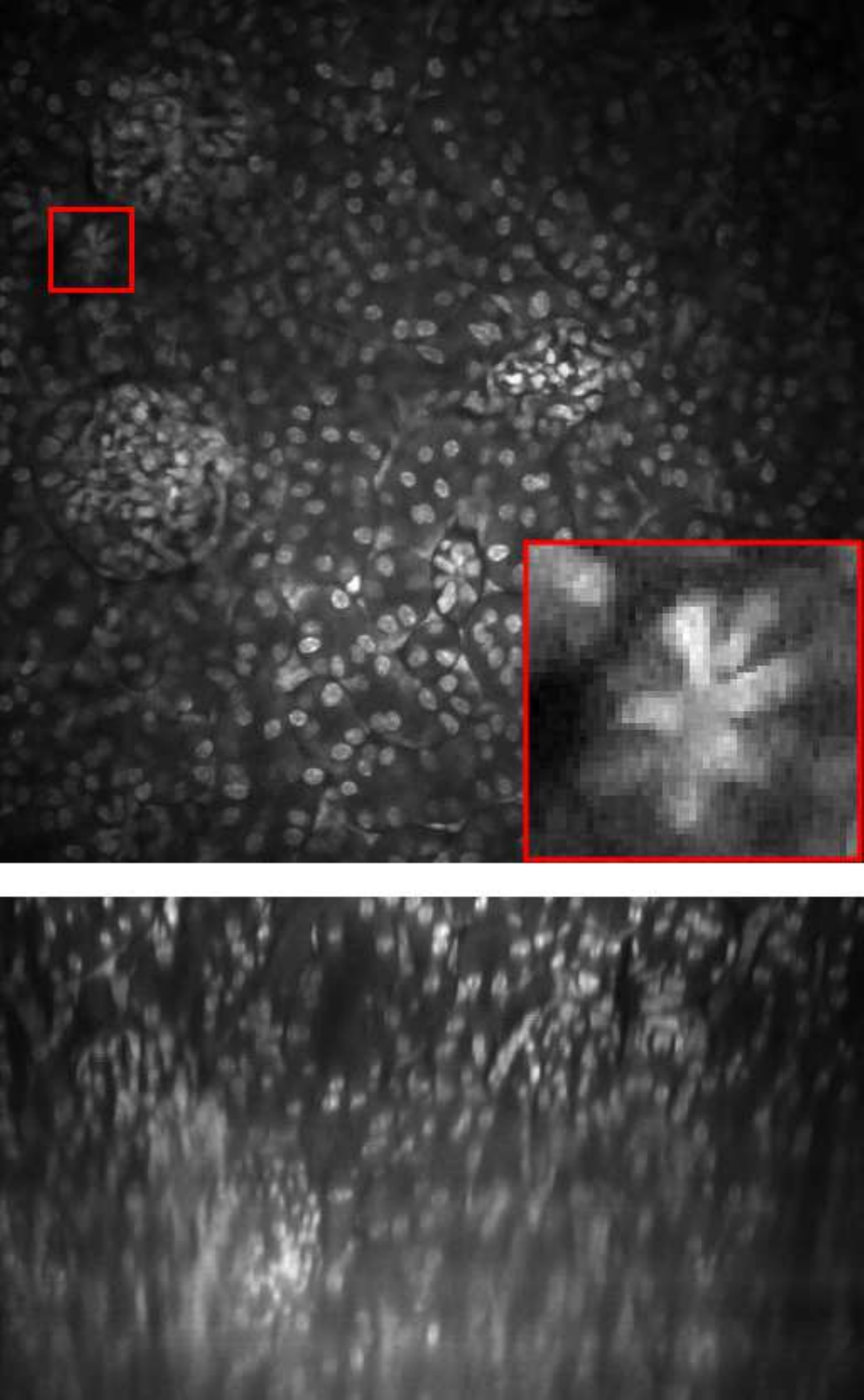}}
	\,
	\subfloat[]
		 {\label{fig:WSM_B_iterVSTpoissonDeb}\includegraphics[width=0.14\textwidth]{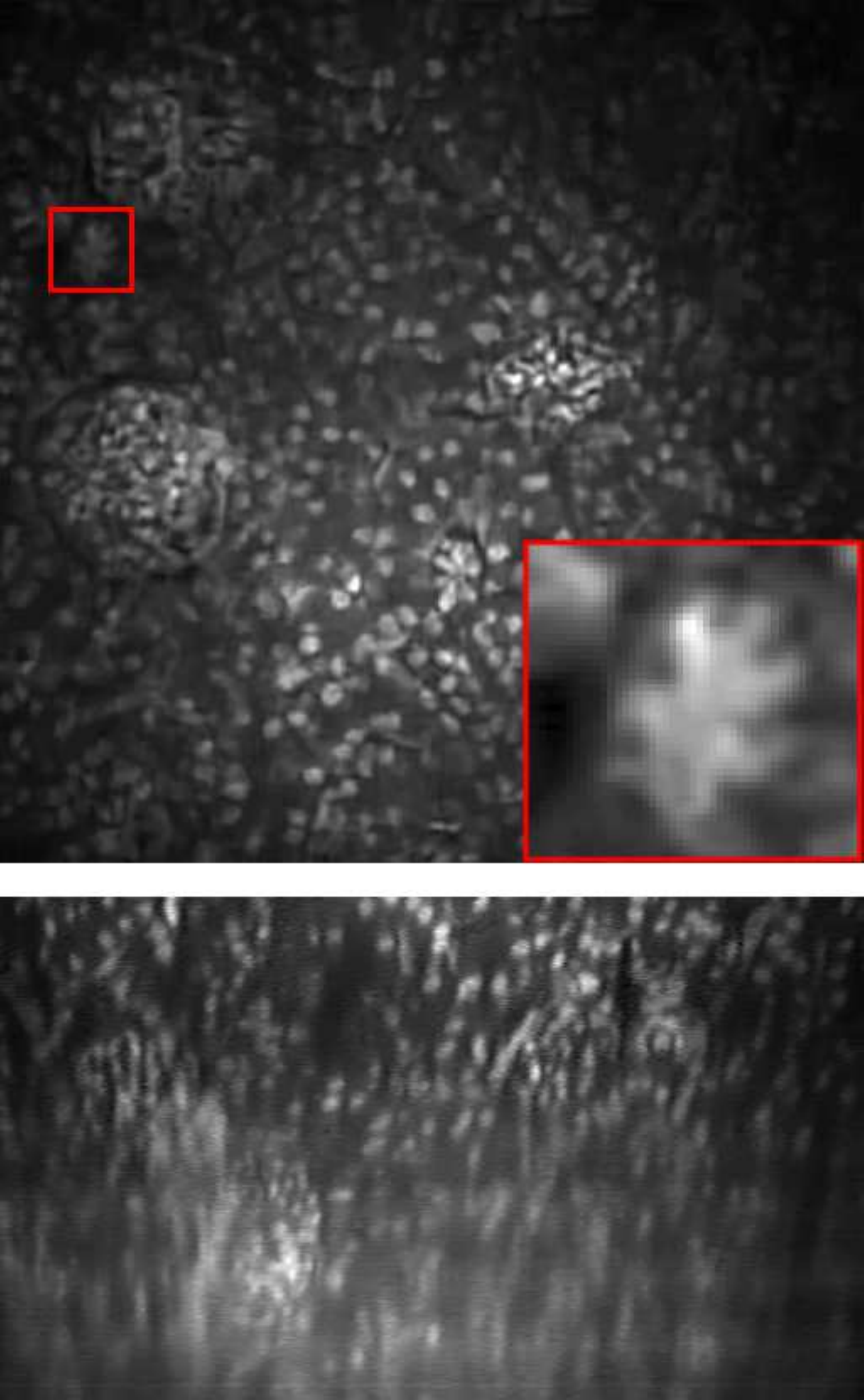}}
	\,
	\subfloat[]
		 {\label{fig:WSM_B_3DacIC}\includegraphics[width=0.14\textwidth]{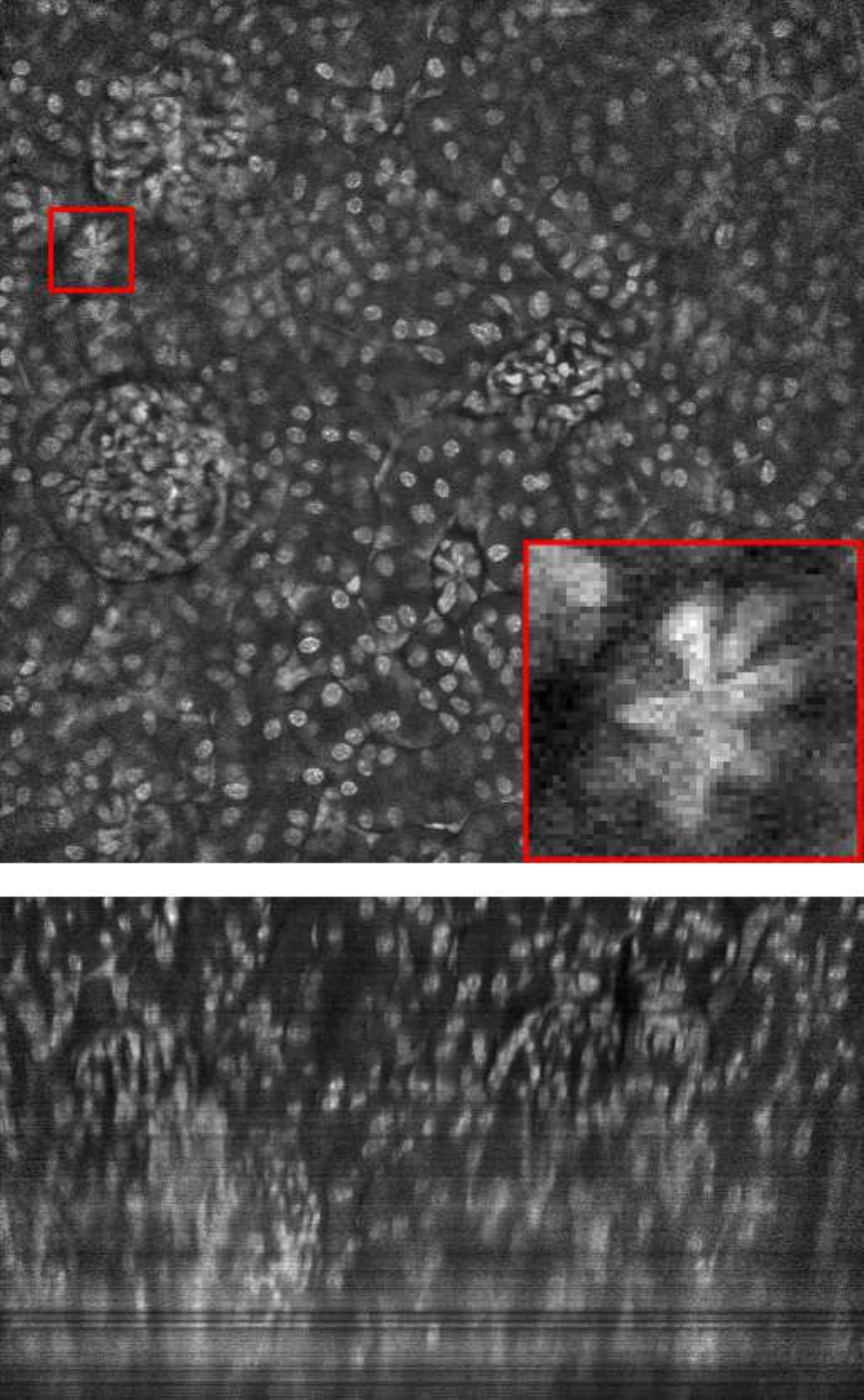}}
	\,
	\subfloat[]
		 {\label{fig:WSM_B_SpCycleGAN3ways}\includegraphics[width=0.14\textwidth]{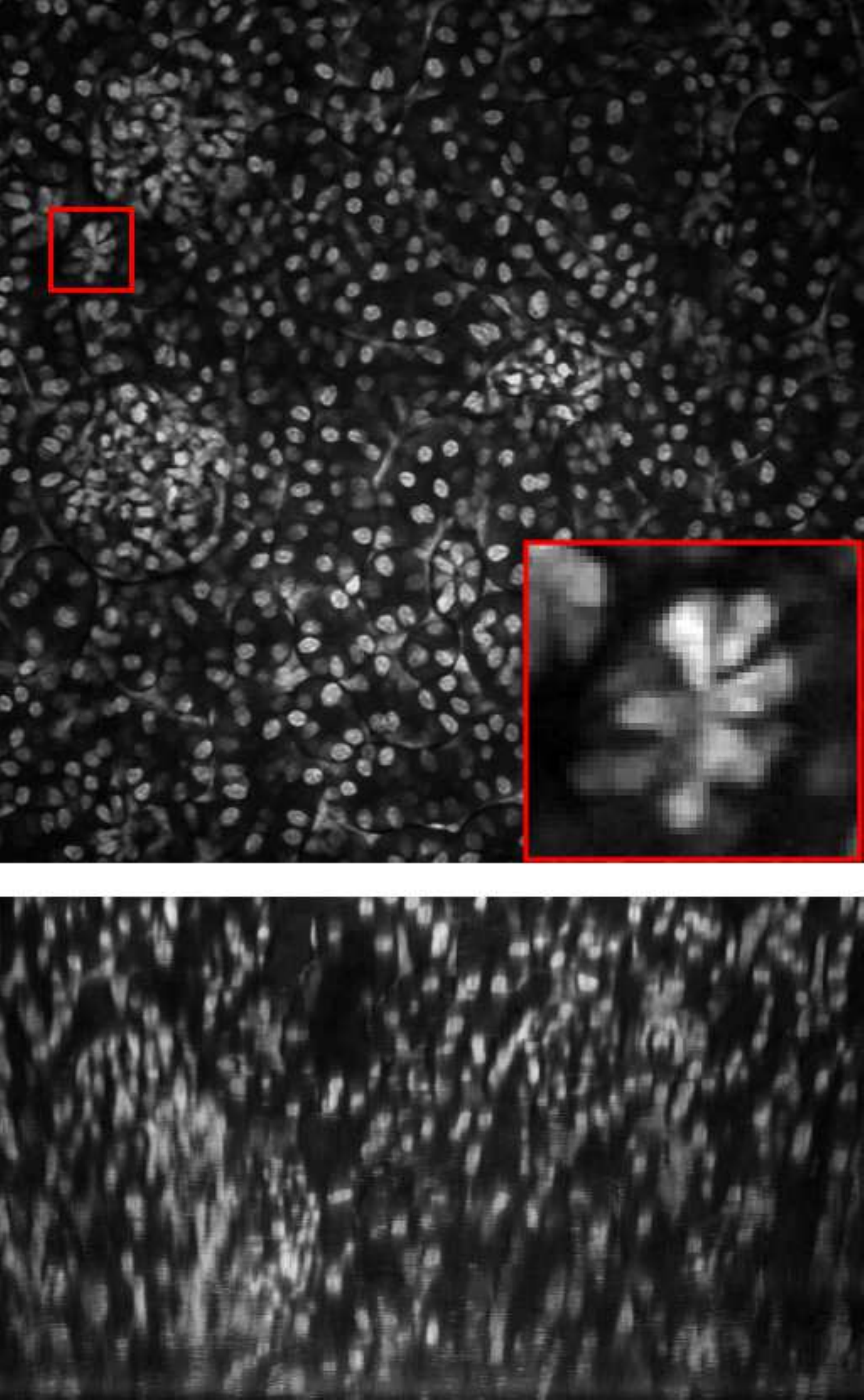}} \\
	 \vspace{-0.1in}
 	\subfloat[]
		 {\label{fig:WSM_R_testA}\includegraphics[width=0.14\textwidth]{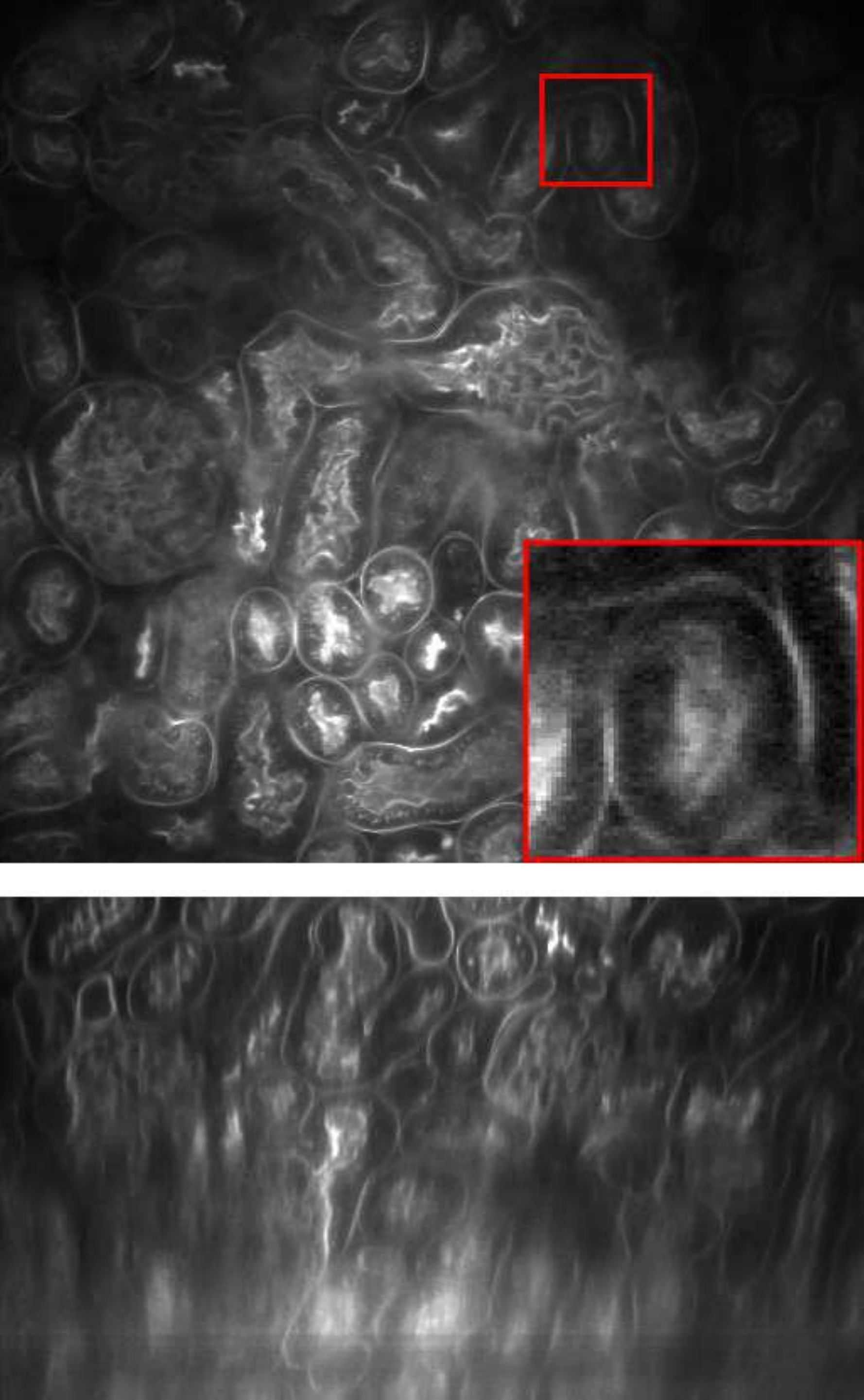}}
	\,
	\subfloat[]
		 {\label{fig:WSM_R_PSFSynRL}\includegraphics[width=0.14\textwidth]{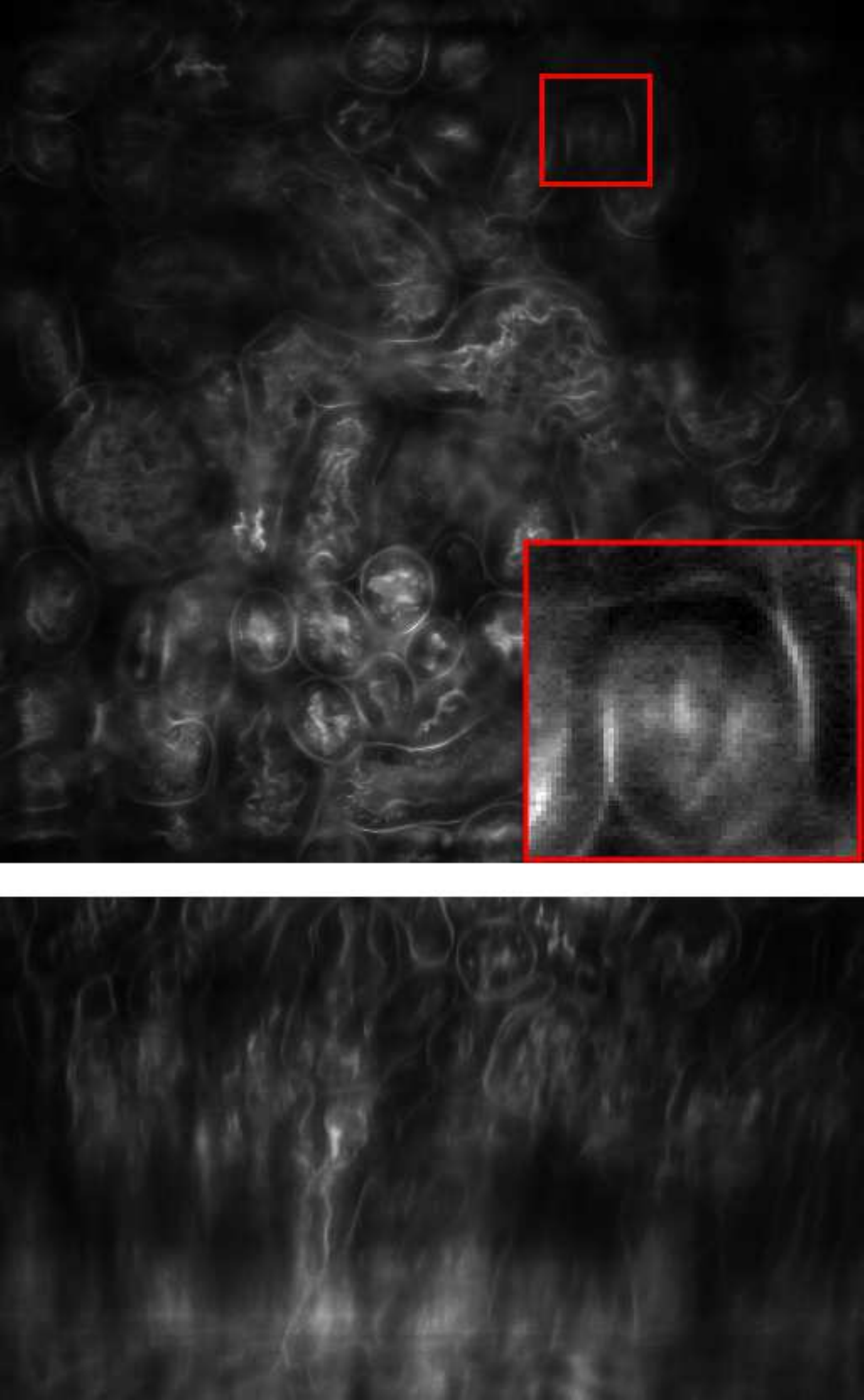}}
	\,
	\subfloat[]
		 {\label{fig:WSM_R_EpiDEMIC}\includegraphics[width=0.14\textwidth]{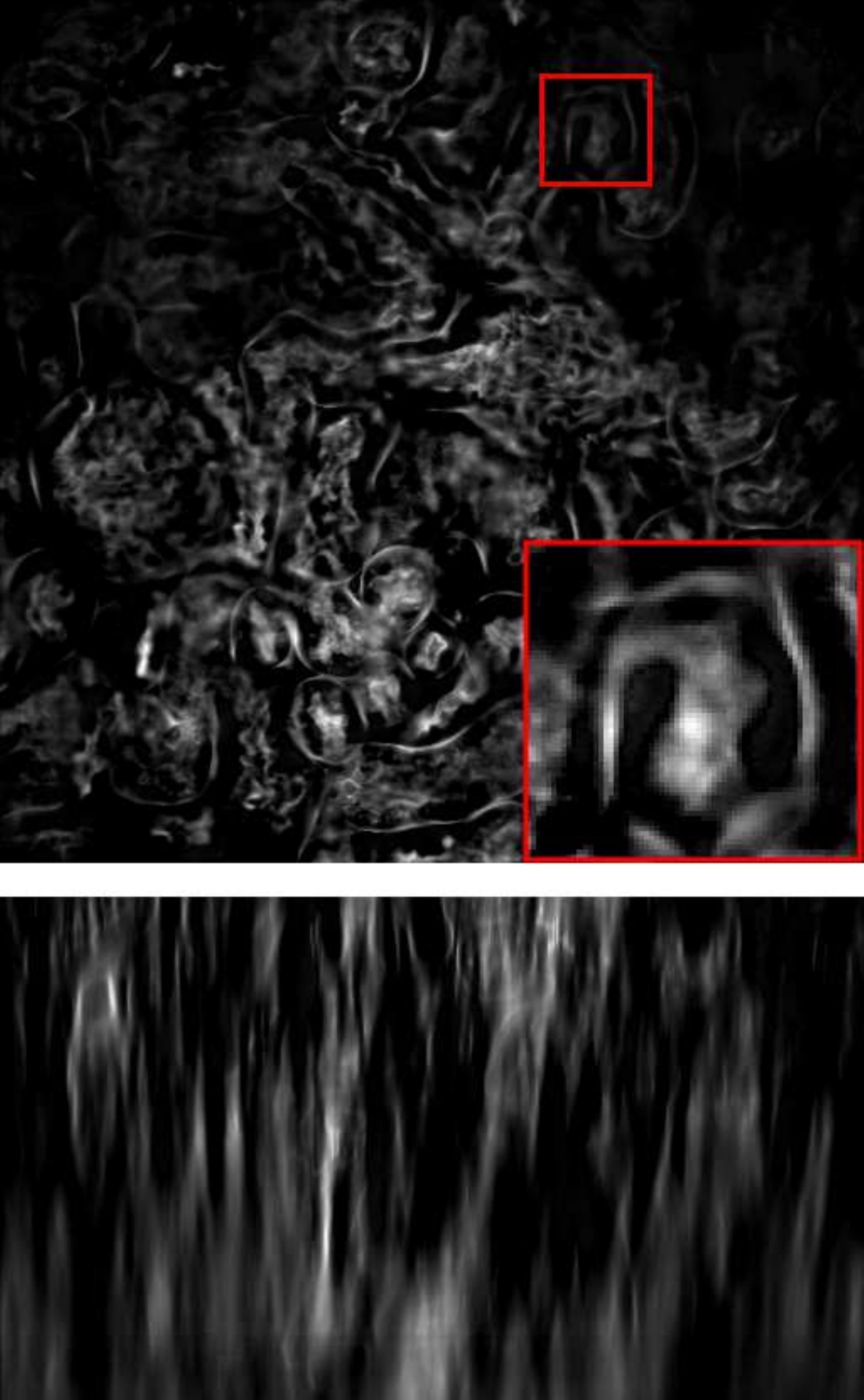}}
	\,
	\subfloat[]
		 {\label{fig:WSM_R_PureDenoise}\includegraphics[width=0.14\textwidth]{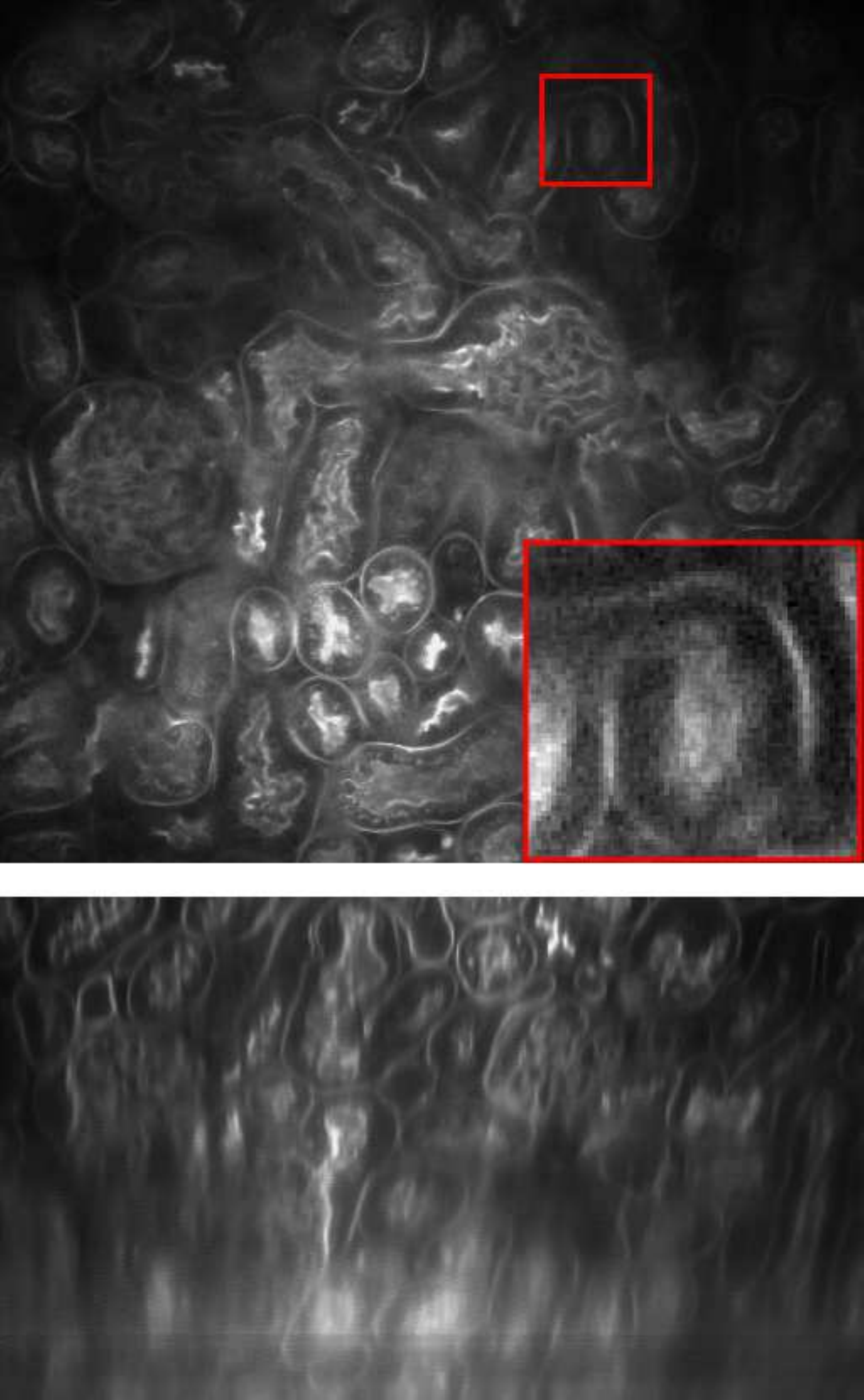}}
	\,
	\subfloat[]
		 {\label{fig:WSM_R_iterVSTpoissonDeb}\includegraphics[width=0.14\textwidth]{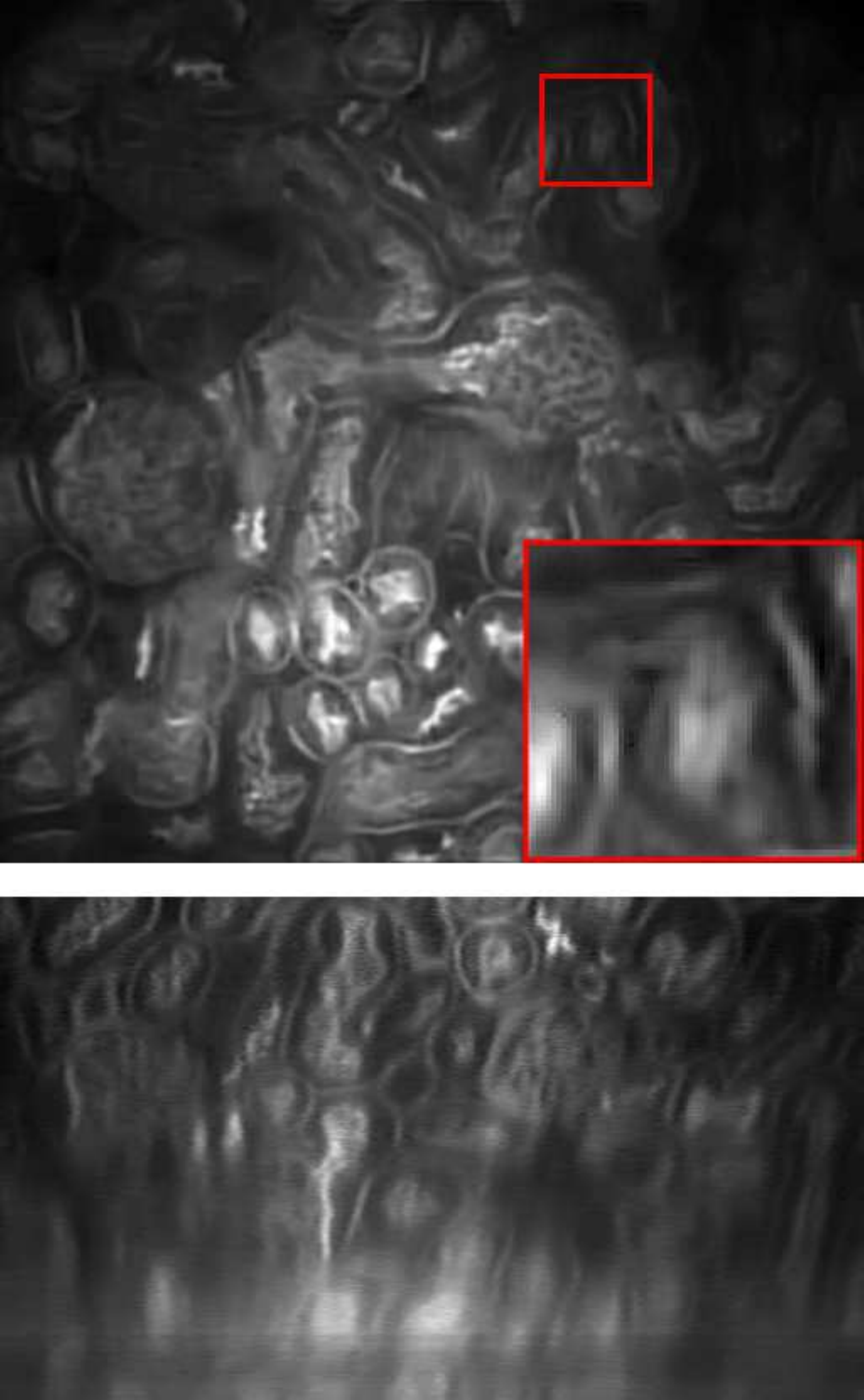}}
	\,
	\subfloat[]
		 {\label{fig:WSM_R_3DacIC}\includegraphics[width=0.14\textwidth]{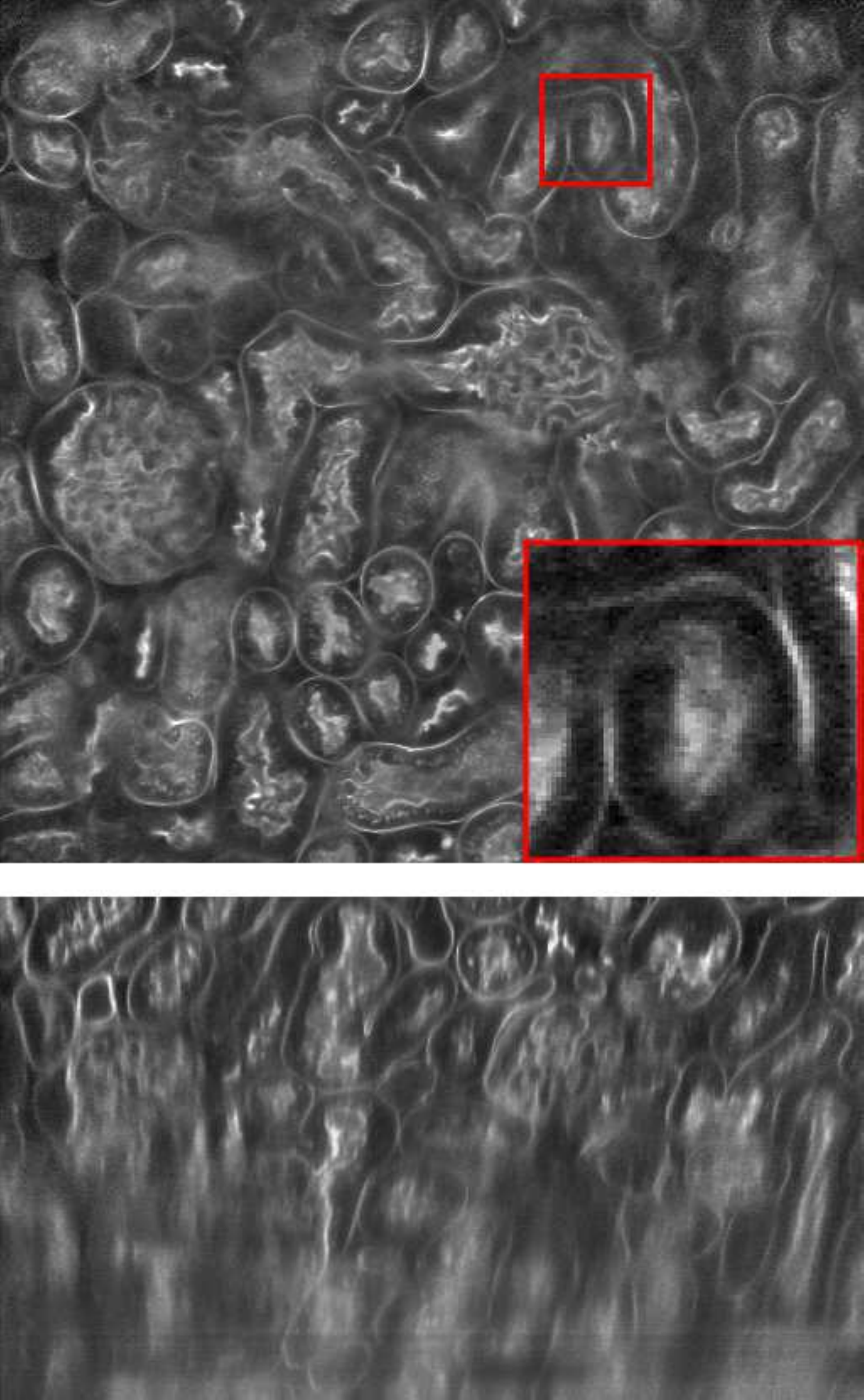}}
	\,
	\subfloat[]
		 {\label{fig:WSM_R_SpCycleGAN3ways}\includegraphics[width=0.14\textwidth]{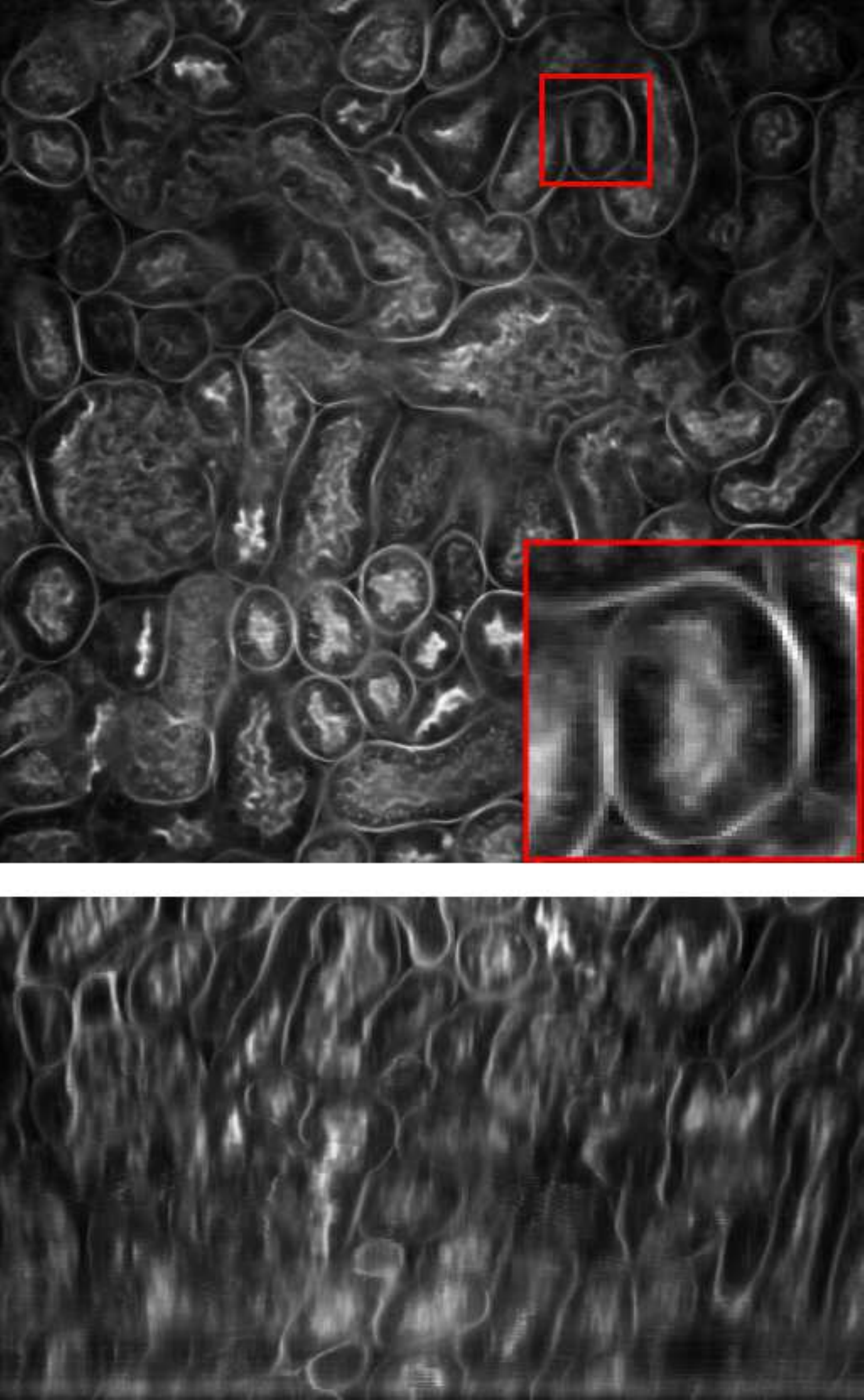}} \\
	\vspace{-0.1in}
\caption{Comparison of the original volume and 3D restored volume results with the $xy$ and $xz$ sections of \textit{Dataset-I} (up) and \textit{Dataset-II} (down) using various methods. First column: Test volume $xy$ section ($I^{OT}_{z126}$) and $yz$ section ($I^{OT}_{y256}$), Second column: RL \cite{bib:Richard1972, bib:Lucy1974}, Third column: EpiDEMIC \cite{bib:Soulez2014}, Fourth column: PureDenoise \cite{bib:PureDenoise}, Fifth column: iterVSTpoissonDeb \cite{bib:Azzari2017}, Sixth column: 3DacIC \cite{bib:SLee2017}, Seventh column: 3-Way SpCycleGAN (Proposed)}
\label{fig:visual_comparison1} 
\vspace{-0.25in}
\end{figure*}

\vspace{-0.05in}
\subsection{3-Way SpCycleGAN and Volumes Averaging}
\vspace{-0.05in}
One drawback of the SpCycleGAN is that it works only in 2D. 
Since our fluorescence microscopy data is a 3D volume, we form the 3D volume by stacking 2D images obtained from different focal planes during data acquisition \cite{bib:DeconvolutionLab2}.
Therefore, we employ $3$-Way SpCycleGAN training which uses the SpCycleGAN in the $xy$, $xz$, and $yz$ sections independently and obtain the generative models per each sections. We use three generative models ($G_{AB}^{xy}$, $G_{AB}^{xz}$, and $G_{AB}^{yz}$) for the inference using $I^{OT}$ which transfer noisy and out-of-focus images to well-defined and focused images in the $xy$, $xz$, and $yz$ sections. 
More specifically, the test volume is sliced into three sets of sectional images and each image is used as an input of inference to generate synthetic well-defined and focused image. Then, these synthetically generated images are stacked in the $z$-, $y$-, and $x$-direction to form $I^{STxy}$, $I^{STxz}$, and $I^{STyz}$, respectively. In general, the number of the $xy$, $xz$, and $yz$ sections are different from each other, we use zero padding to make the dimension of three volumes identical. 
Lastly, the final volume ($I^F$) is obtained as
\begin{equation}
I^F = w_1 I^{STxy} + w_2 I^{STxz} + w_3 I^{STyz} 
\end{equation}
where $w_1$, $w_2$, and $w_3$ are weight coefficients of $I^{STxy}$, $I^{STxz}$, and $I^{STyz}$, respectively.

\section{EXPERIMENTAL RESULTS}
\label{sec:experimental_results}
\vspace{-0.05in}

The performance of our proposed deconvolution method was tested on two different datasets:\footnote{\textit{Dataset-I} and \textit{II} were provided by Malgorzata Kamocka of the Indiana Center for Biological Microscopy.} \textit{Dataset-I} and \textit{II}. \textit{Dataset-I} and \textit{II} are originally obtained at the same time with different fluorophores to delineate different biological structures. \textit{Dataset-I} and \textit{II} are both comprised of $Z = 512$ grayscale images, each of size $X \times Y = 512 \times 512$ pixels. We selected a blurred and noisy subvolume ($I^{OA}$) from last $200$ images of given fluorescence microscopy volume as $I^{O}_{(1:512,1:512,313:512)}$ with a size of $512 \times 512 \times 200$ for \textit{Dataset-I} and \textit{II}. Also, a good quality subvolume ($I^{OB}$) was selected based on a biologist's opinion as $I^{O}_{(1:512,1:512,15:214)}$ with size of $512 \times 512 \times 200$ for \textit{Dataset-I} and \textit{II}. The test volume ($I^{OT}$) for each dataset was selected at deeper tissue depth than $I^{OB}$ as $I_{(1:512,1:512,215:512)}^{O}$ with size of $512 \times 512 \times 298$. 

Our $3$-Way SpCycleGAN is implemented in PyTorch using the Adam optimizer \cite{bib:Kingma2014} with constant learning rate $0.0002$ for the first $100$ epochs and gradually decreased to $0$ for the next $100$ epochs. Also, we use the ResNet $9$ blocks \cite{bib:ResNet} for all generative models ($G_{AB}$, $G_{BA}$, and $H$) with $64$ feature maps at the first layer. For the corresponding discriminative models ($D_A$ and $D_B$), same discriminative models are used in the CycleGAN \cite{bib:CycleGAN}. 
We randomly select patches size of $256 \times 256$ from $512 \times 512$ for the $xy$ sections and $200 \times 200$ from $512 \times 200$ for the $xz$ and $yz$ sections for the SpCycleGAN training, respectively. We choose larger resolution for the $xy$ sections since $xy$ sections is a finer resolution than those $xz$ and $yz$ sections. 
Also, we set the coefficients $\lambda_1 = \lambda_2 = 10$ for all $3$-Way SpCycleGAN training for both \textit{Dataset-I} and \textit{II}. 
Lastly, the weights for $3$-way volume averaging is set as $w_1 = w_2 = w_3 = 1/3$ so that each sectional results equally contribute the final volume. 

\begin{table*}[htb!]
\centering
\renewcommand{\tabcolsep}{3pt}
{\begin{tabular}{|c||c|c|c||c|c|c|}
\hline
& \multicolumn{3}{c||}{$I^{OT}_{(1:512,1:512,149:298)}$ of the \textit{Dataset-I}} & \multicolumn{3}{c|}{$I^{OT}_{(1:512,1:512,149:298)}$ of the \textit{Dataset-II}} \\
\hline
\multirow{2}{*}{Method} & 3-Way  & 3-Way & 3-Way  & 3-Way  & 3-Way  & 3-Way  \\
& BRISQUE \cite{bib:BRISQUE} & OG-IQA \cite{bib:OGIQA} & Microscopy IFQ \cite{bib:Yang2018} & BRISQUE \cite{bib:BRISQUE} & OG-IQA \cite{bib:OGIQA} & Microscopy IFQ \cite{bib:Yang2018} \\
\hline
\textit{$I^{OT}_{(1:512,1:512,149:298)}$} & $35.50$ & $-0.34$ & $1.95$ & $\mathbf{15.68}$ & $-0.64$ & $3.07$ \\
\hline
RL \cite{bib:Richard1972, bib:Lucy1974} & $41.19$ & $-0.80$ & $0.67$ & $23.97$ & $-0.49$ & $3.98$ \\
\hline
EpiDEMIC \cite{bib:Soulez2014} & $58.96$ & $-0.75$ & $0.62$ & $50.97$ & $-0.29$ & $0.96$ \\
\hline
PureDenoise \cite{bib:PureDenoise} & $39.90$ & $-0.67$ & $2.04$ & $24.40$ & $-0.47$ & $3.34$ \\
\hline
iterVSTpoissonDeb \cite{bib:Azzari2017} & $35.01$ & $-0.44$ & $2.84$ & $32.38$ & $-0.36$ & $4.05$ \\
\hline
3DacIC \cite{bib:SLee2017} & $37.96$ & $-0.26$ & $0.66$ & $19.68$ & $-0.61$ & $1.64$ \\
\hline
3-Way SpCycleGAN & \multirow{2}{*}{$\mathbf{34.05}$} & \multirow{2}{*}{$\mathbf{-0.88}$} & \multirow{2}{*}{$\mathbf{0.52}$} & \multirow{2}{*}{$31.14$} & \multirow{2}{*}{$\mathbf{-0.82}$} & \multirow{2}{*}{$\mathbf{0.94}$} \\
(Proposed) &   &   &   &   &   &  \\
\hline
\end{tabular}
}
\vspace{-0.1in}
\caption{Comparison of the performance of proposed and other restoration methods with three image quality metrics using \textit{Dataset-I} and \textit{II}}
\label{tab:results_comp_number}
\vspace{-0.25in}
\end{table*}

Our proposed deconvolution results were visually compared with five different techniques including RL \cite{bib:Richard1972, bib:Lucy1974}, EpiDEMIC \cite{bib:Soulez2014}, PureDenoise \cite{bib:PureDenoise}, iterVSTPoissonDeb \cite{bib:Azzari2017}, and 3DacIC \cite{bib:SLee2017} shown in Figure \ref{fig:visual_comparison1}. Note that we used default settings for the methods RL and PureDenoise in ImageJ plugins, EpiDEMIC in Icy plugin, and iterVSTpoissonDeb. 

As shown in Figure \ref{fig:visual_comparison1}, first column displays a sample $xy$ section ($I^{OT}_{z126}$) and $xz$ section ($I^{OT}_{y256}$) of original test volumes in \textit{Dataset-I} and \textit{II}, respectively. 
The original test volumes suffer from significant intensity inhomogeneity, blur, and noise. Also, this degradation gets worse at deeper depth as shown in the $xz$ section. 
As observed, our proposed method showed the best performance among presented methods in terms of inhomogeneity correction, clarity of the shape of nuclei and tubules/glomeruli structure, and noise level. 
More specifically, two deconvolution methods (RL and EpiDEMIC) successfully reduced blur but the original shapes of the biological structures were lost. 
Also, EpiDEMIC's $xz$ section deconvolution results were all connected each other since EpiDEMIC learned 2D features as a prior to enhance 3D. 
Similarly, two denoising methods (PureDenoise and iterVSTpoissonDeb) successfully suppressed Poisson noise but these denoising results were still suffered from intensity inhomogeneity and blur. In fact, the denoising results added more blur than original test volume. 
Meanwhile, 3DacIC method successfully corrected inhomogeneity but this method amplified background noise level and aggravated image quality. Moreover, 3DacIC exacerbated line shape noise shown in the $xz$ sections. 

In addition to the visual comparison, three image quality metrics were utilized for evaluating volume quality of restored volumes of proposed and other presented methods. Since our microscopy volumes do not have reference volumes to compare, we need to use no reference image quality assessment (NR-IQA) \cite{bib:BRISQUE} instead of traditional PSNR, SSIM, and FSIM. 
One problem is that there is no gold standard image quality metric for 3D fluorescence microscopy. We employed the blind/referenceless image spatial quality evaluator (BRISQUE) \cite{bib:BRISQUE}, the oriented-gradient image quality assessment (OG-IQA) \cite{bib:OGIQA}, and the microscopy image focus quality assessment (Microscopy IFQ) \cite{bib:Yang2018} for evaluating quality of restored microscopy volumes. 
In particular, BRISQUE model is a regression model learned from local statistics of natural scenes in the spatial domain to measure image quality where the quality value range is from $0$ to $100$. 
OG-IQA model is a gradient feature based model that maps from image features to image quality via an adaboosting back propagation neural network where the quality value is from $-1$ to $1$. 
Lastly, Microscopy IFQ measures discrete defocus level from $0$ to $10$ using $84 \times 84$ local patches using a CNN. 
Instead of using discrete defocus level, we got a probability ($p(l)$) for each corresponding defocus level ($l$) before the softmax layer and used these probabilities and corresponding defocus levels to obtain expected value which defined as
\vspace{-0.1in}
\begin{equation}
\text{Microscopy IFQ} = \sum_{l=0}^{10} l \cdot p(l).
\vspace{-0.05in}
\end{equation}
Since this Microscopy IFQ value was obtained for each individual $84 \times 84$ local patches, we resized our images to be nearest integer multiple of local patch size and took an average through entire image. Note that the smaller values of all three image quality assessments indicate the better image quality. In addition, since these three image quality assessments can only measure the quality of 2D images, we again utilized $3$-way idea to obtain the image quality of the $xy$, $xz$, and $yz$ sections and took an average of them. The end result was a single representative value for volume quality per each volume.

We used these three image quality metrics to test seven different volumes including the original test volume.
This is provided in Table \ref{tab:results_comp_number}. Note that we selected the $150$ most blurred and noisy image volumes from test volume as $I^{OT}_{(1:512,1:512,149:298)}$ for the evaluation purpose. As mentioned above the smaller values are considered to be indicators of the better image volume quality. As observed in Table \ref{tab:results_comp_number}, our proposed method outperformed the other methods and original volume except from $3$-Way BRISQUE in \textit{Dataset-II}. 
This is because BRISQUE measures the quality from natural image statistics and this model is a favor of blurred volume. 
Therefore, RL and EpiDEMIC had higher values in BRISQUE image quality metric. 
Similarly, OG-IQA is a gradient based measurement so edge preserved restoration volume can get smaller image quality values. 
Also, Microscopy IFQ is a defocus level measurement. Hence, PureDenoise and iterVSTpoissonDeb had sometimes higher OG-IQA and Microscopy IFQ. 
3DacIC produced reasonably lower values for the $xy$ sections but the quality of the $xz$ and $yz$ sections were poor so that the entire volume quality was inferior than proposed method's volume quality.

\begin{figure}[htb!]
	\centering
	\subfloat[]
		 {\label{fig:WSM_B_testA}\includegraphics[width=0.15\textwidth]{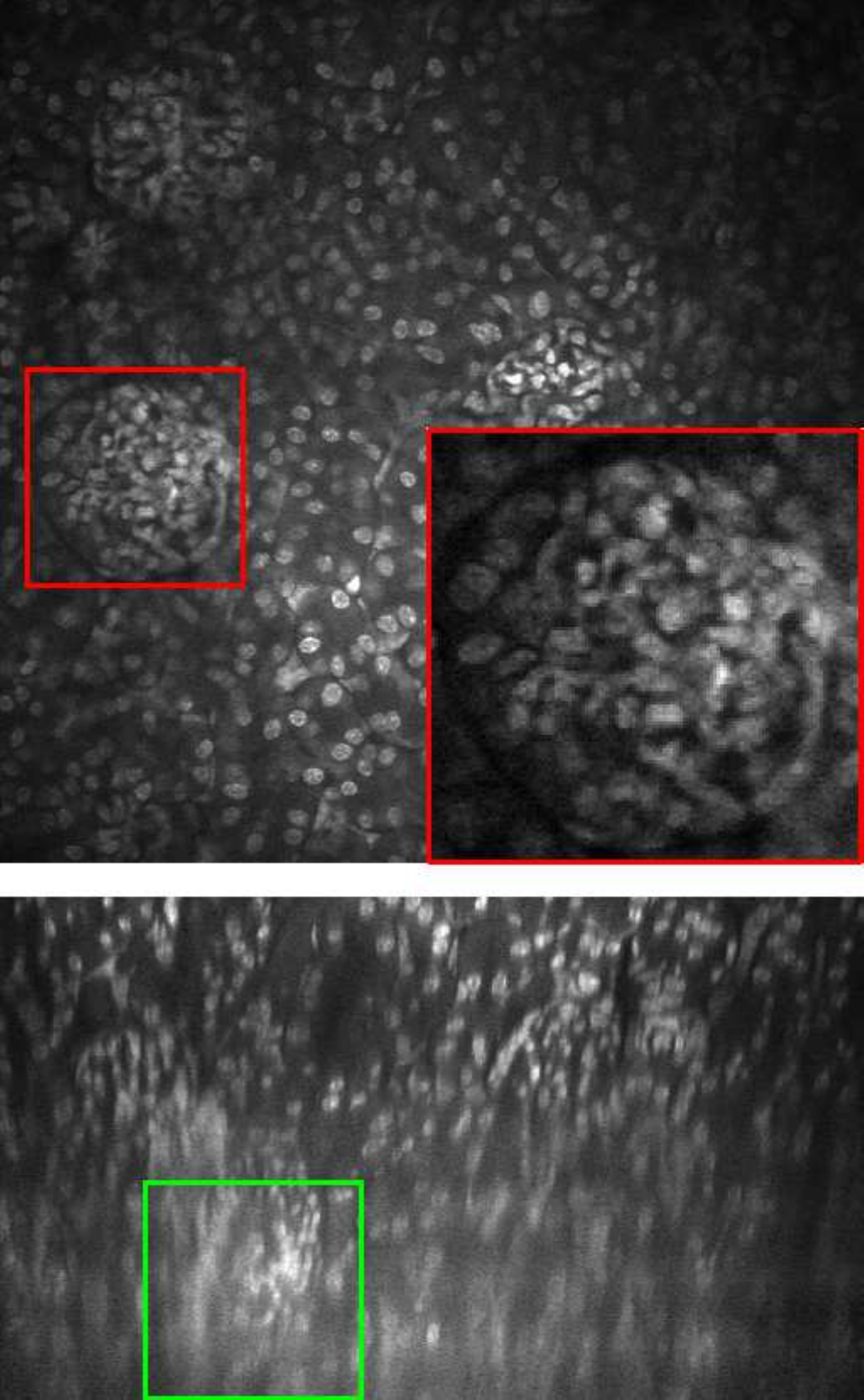}}
	\,\,
	\subfloat[]
		 {\label{fig:WSM_B_SpCycleGAN3ways}\includegraphics[width=0.15\textwidth]{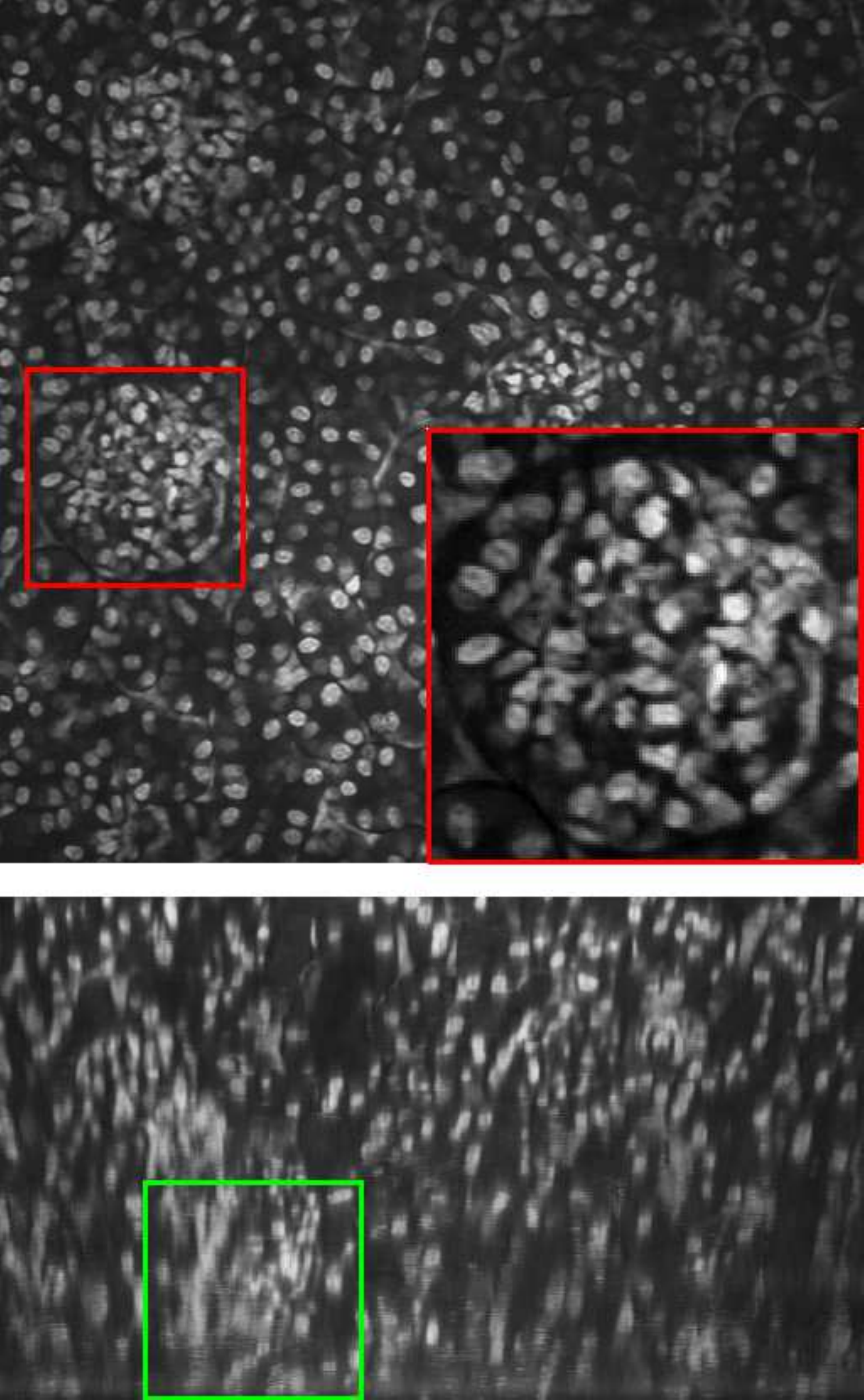}}
	\,\,
	\subfloat[]
		 {\label{fig:WSM_B_SpCycleGANxy}\includegraphics[width=0.15\textwidth]{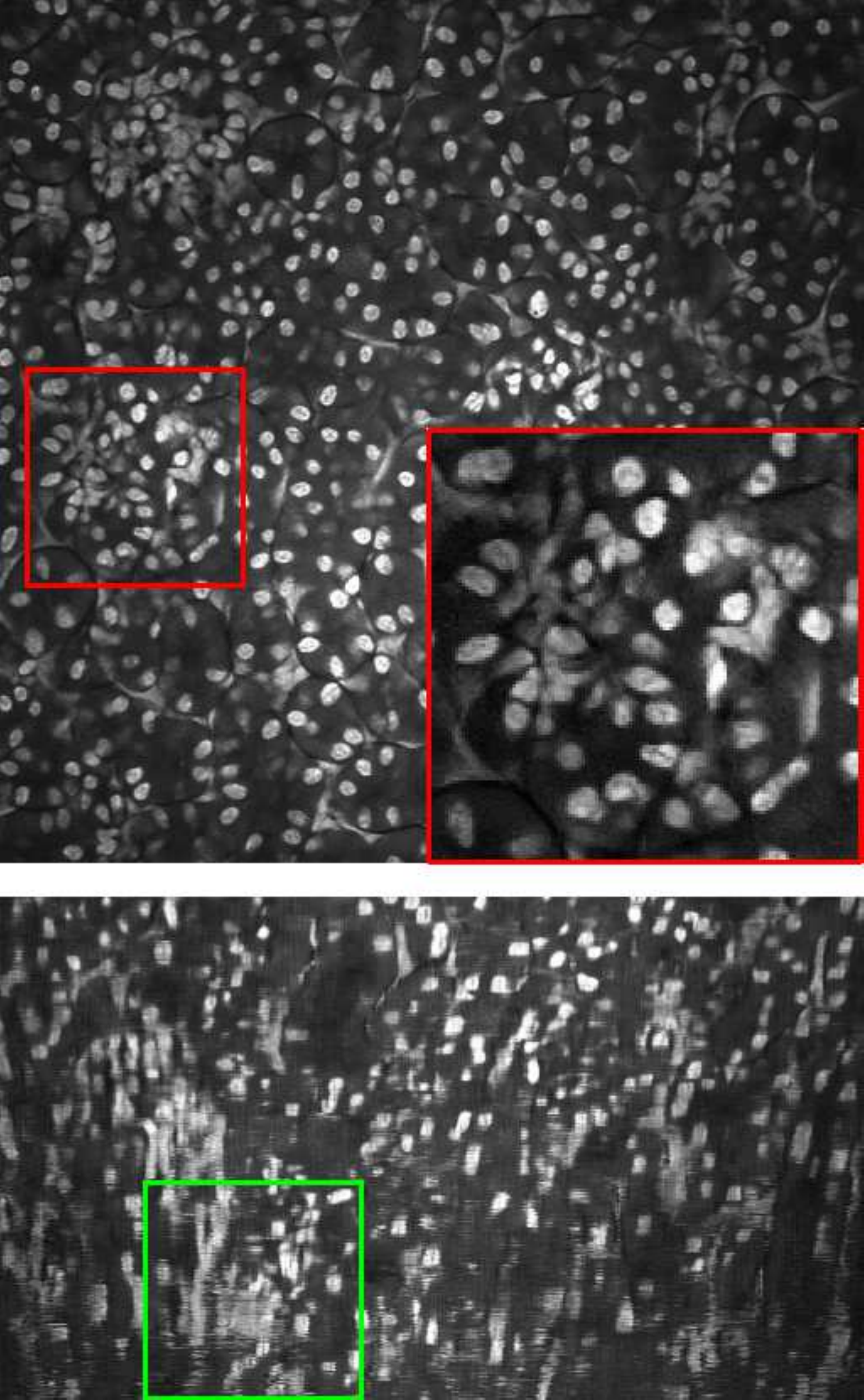}}
	 \vspace{-0.15in}
\caption{Comparison of original test volume $xy$ section ($I^{OT}_{z126}$) and $yz$ section ($I^{OT}_{y256}$), restored volume using proposed $3$-Way SpCycleGAN, and restored volume using the $xy$ sections of SpCycleGAN using \textit{Dataset-I}}
\label{fig:visual_comparison2} 
\vspace{-0.2in}
\end{figure}

Lastly, Figure \ref{fig:visual_comparison2} portrays the visual comparison between proposed $3$-Way SpCycleGAN and SpCycleGAN using the $xy$ sections only ($w_1 = 1$, $w_2 = w_3 = 0$). Without having $z$-direction information, SpCycleGAN using the $xy$ sections cannot correctly restore the glomerulus displayed in the red box. Also, the $z$-direction images are frequently discontinued as shown in the $xz$ section in the green box. Compared to that, proposed $3$-Way SpCycleGAN can successfully restore glomerulus and connect smoothly in $z$-direction.

\vspace{-0.05in}
\section{CONCLUSION AND FUTURE WORK}
\label{sec:conclusion}
\vspace{-0.1in}
This paper has presented a blind image deconvolution method for fluorescence microscopy volumes using the $3$-Way SpCycleGAN. 
Using the $3$-Way SpCycleGAN, we can successfully restore the blurred and noisy volume to good quality volume so that deeper volume can be used for the biological research. Future work will include developing a 3D segmentation technique using our proposed deconvolution method as a preprocessing step.
\vspace{-0.05in}
\bibliographystyle{IEEEbib}
\bibliography{ISBI2019_bib}

\end{document}